\documentclass[runningheads]{llncs}
\usepackage{graphicx}

\usepackage[table]{xcolor}
\usepackage{tikz}
\usepackage{comment}
\usepackage{amsmath,amssymb} 
\usepackage{color}
\usepackage{wrapfig}

\usepackage[accsupp]{axessibility}  

\usepackage[pagebackref,breaklinks,colorlinks,citecolor=blue]{hyperref}
\usepackage{xspace}
\usepackage[shortlabels]{enumitem}
\usepackage{booktabs}
\usepackage[outercaption]{sidecap}
\usepackage{pifont}
\newcommand{\cmark}{\ding{51}}%
\newcommand{\xmark}{\ding{55}}%

\usepackage{amssymb}
\usepackage{amsmath,amsfonts}
\usepackage{amsopn}
\usepackage{bm} 
\usepackage{multirow}
\usepackage{threeparttable}
\usepackage{tabularx}
\usepackage{tablestyles}
\usepackage{arydshln}
\usepackage{float}

\newcommand{\ProbOpr}[1]{\mathbb{#1}}

\newcommand{\expect}[2]{%
\ifthenelse{\equal{#2}{}}{\ProbOpr{E}_{#1}}
{\ifthenelse{\equal{#1}{}}{\ProbOpr{E}\left[#2\right]}{\ProbOpr{E}_{#1}\left[#2\right]}}} 
\newcommand{\var}[2]{%
\ifthenelse{\equal{#2}{}}{\ProbOpr{VAR}_{#1}}
{\ifthenelse{\equal{#1}{}}{\ProbOpr{VAR}\left[#2\right]}{\ProbOpr{VAR}_{#1}\left[#2\right]}}} 

\newcommand{\eat}[1]{}

\newcommand{\ourmethod}{{\sc {FreeSeg}}\xspace}
\newcommand{\ourmethodbf}{{\textsc{\textbf{FreeSeg}}}\xspace}
\newcommand\mypara[1]{\vskip 2pt \noindent\textbf{#1}}

\definecolor{LightCyan}{rgb}{0.88,1,1}

\newcommand{\eg}{{\em e.g.}}
\newcommand{\ie}{{\em i.e.}}
\newcommand{\etal}{{\em et al.}}
\usepackage[capitalize]{cleveref}
\crefname{section}{Sec.}{Secs.}
\Crefname{section}{Section}{Sections}
\Crefname{table}{Table}{Tables}
\crefname{table}{Tab.}{Tabs.}

\begin{document}
\pagestyle{headings}
\mainmatter
\def\ECCVSubNumber{****}  

\title{Learning with Free Object Segments for Long-Tailed Instance Segmentation} 


\titlerunning{Learning with Free Object Segments for Long-Tailed Instance Segmentation}
\author{Cheng Zhang \and
Tai-Yu Pan \and
Tianle Chen \and \\
Jike Zhong \and
Wenjin Fu \and 
Wei-Lun Chao
}
\authorrunning{C. Zhang \etal}
%
\institute{The Ohio State University, Columbus OH 43210, USA \\
}
\maketitle

\begin{abstract}
One fundamental challenge in building an instance segmentation model for a large number of classes in complex scenes is the lack of training examples, especially for rare objects.
In this paper, we explore the possibility to increase the training examples without laborious data collection and annotation.
We find that an abundance of instance segments can potentially be obtained freely from object-centric images, according to two insights: (i) an object-centric image usually contains one salient object in a simple background; (ii) objects from the same class often share similar appearances or similar contrasts to the background. Motivated by these insights, we propose a simple and scalable framework \ourmethod for extracting and leveraging these ``free'' object foreground segments to facilitate model training in long-tailed instance segmentation. 
Concretely, we investigate the similarity among object-centric images of the same class to propose candidate segments of foreground instances, followed by a novel ranking of segment quality. 
The resulting high-quality object segments can then be used to augment the existing long-tailed datasets, \eg, by copying and pasting the segments onto the original training images. Extensive experiments show that \ourmethod yields substantial improvements on top of strong baselines and achieves state-of-the-art accuracy for segmenting rare object categories. 
\end{abstract} 

\section{Introduction}
\label{s_intro}

\begin{figure}[t]
    \setlength{\abovecaptionskip}{1mm}
    \setlength{\belowcaptionskip}{-5mm}
    \centerline{\includegraphics[width=1\linewidth]{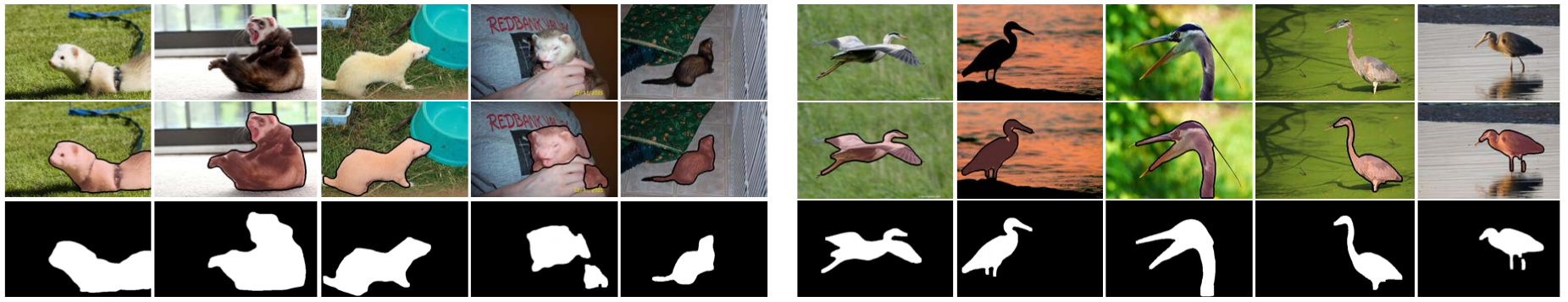}}
    \caption{\small \textbf{Illustration of our approach \ourmethod.} We sample two rare classes, \textit{ferret} (left) and \textit{heron} (right) from LVIS v1~\cite{gupta2019lvis}, and retrieve 
    object-centric images (the upper row of each class) from the ImageNet dataset~\cite{russakovsky2015imagenet}. We then show the discovered object segments (middle) and binary masks (bottom) by \ourmethod. The abundant object segments have diverse appearances and poses and can be effectively used to improve the instance segmentation.}
    \label{fig:1}
\end{figure}

Object detection and instance segmentation are fundamental building blocks for many high-impact real-world applications (\eg, autonomous driving).
Recent years have witnessed an unprecedented breakthrough in both of them, thanks to deep neural networks~\cite{ren2016faster,he2017mask,girshick2014rich} and large-scale datasets for common objects (\eg, persons and cars)~\cite{lin2014microsoft,everingham2010pascal,zou2019object}. Yet, when it comes to rare, less commonly seen objects (\eg, an unusual traffic sign), 
there is a drastic performance drop due to insufficient training examples~\cite{gupta2019lvis,shao2019objects365,zhang2021deep}. 
This challenge has attracted significant attention lately in how to learn an object detection or instance segmentation model given labeled data of a ``long-tailed'' distribution across classes~\cite{zhu2014capturing}. Specifically, a number of works have been dedicated to developing new training algorithms, objectives, or model architectures~\cite{oksuz2020imbalance,hu2020learning,li2020overcoming,wang2020frustratingly,tan2020equalization,wu2020forest,gupta2019lvis,wang2020seesaw,tan2020equalizationv2,wang2020devil}.

In this paper, we explore a drastically different approach.
\emph{We investigate the possibility of obtaining more labeled instances (\ie, instance segments of objects) at a minimal cost, especially for rare objects.} 
We build upon the recent observation in~\cite{zhang2021simple}
--- many objects do not appear frequently enough in complex scenes but are found frequently alone in object-centric images
--- to acquire an abundance of object-centric images (\eg, ImageNet~\cite{deng2009imagenet} or Google images) for rare classes. 
Zhang~\etal~\cite{zhang2021simple} have shown that, even with only pseudo bounding boxes for these images, they can already improve the detector effectively.

We take one key step further to leverage the underlying properties of object-centric images to create high-quality instance labels that can facilitate both detection and segmentation model training.
In general, object-centric images usually contain one salient object in a relatively simple background than scene-centric images like those in MSCOCO~\cite{lin2014microsoft}. Moreover, objects of the same class usually share similar appearances, shapes, contrasts, or more abstractly, common parts to the background~\cite{rother2006cosegmentation} (see
\autoref{fig:1} for an example). These properties open up the opportunity to discover object segments almost \emph{freely} from object-centric images of the same class --- by exploring their common salient regions.

To this end, we propose a framework named \ourmethod (Free Object Segments) to take advantage of these properties. We first extract the common foreground regions from object-centric images of the same class. This can be done, for example, by off-the-shelf co-segmentation models~\cite{zhang2020deep}.
While not perfect, sometimes missing the true objects or including backgrounds, these extracted regions have surprisingly captured a decent portion of objects with tight segmentation masks. 
{Nevertheless, directly using all of these regions, mixed with false positive and noisy segments, would inevitably introduce a great amount of noise to the downstream tasks.} 
To address this, we propose a novel segment ranking approach to mine the most reliable and high-confident object masks. After all, we aim for a set of high-quality instance segments from object-centric images, not to segment all the object-centric images well. 

{How can we leverage these high-quality instance segments from object-centric images?} One naive way is to directly train the instance segmentation model on the object-centric images, using these segments as supervision. Nevertheless, the fact that these objects mostly show up alone in simple backgrounds makes them somewhat too simple for the model. We, therefore, choose to place these object segments in the context of complex scene-centric images, via simple copy-paste augmentation~\cite{ghiasi2020simple}. Unlike \cite{ghiasi2020simple}, which merely pastes human-annotated segments from one image to another to increase the \emph{context diversity}, our \ourmethod approach brings the best of abundant free object segments to increase the \emph{appearance diversity}, especially for rare object categories. 

We evaluate \ourmethod on the long-tailed LVIS benchmark~\cite{gupta2019lvis}. 
\ourmethod leads to a massive improvement in segmenting rare object instances
by effectively increasing the labeled training data for them. 
Moreover, \ourmethod is detached from the model training phase and is thus model-agnostic. Namely, it can potentially benefit all kinds of instance segmentation model architectures. \ourmethod is also compatible with existing efforts on long-tailed object detection and segmentation to achieve further gains.

In summary, our \ourmethod framework opens up a novel direction that brings the best of discovering pixel-level supervision in object-centric images to facilitate long-tailed instance segmentation. 
Our \textbf{main contributions} are:
\begin{itemize} [itemsep=5pt,topsep=5pt,leftmargin=15pt]
    \item We demonstrate the possibility to increase the number of training examples for instance segmentation without laborious pixel-level data collection and annotation.
    \item We propose a simple and scalable pipeline for discovering, extracting, and leveraging free object foreground segments to facilitate long-tailed instance segmentation.
    \item Our \ourmethod framework shows promising gains on the challenging LVIS dataset and demonstrates a strong compatibility with existing works.
\end{itemize}

\section{Related Work}
\label{s_related}

\mypara{Long-tailed object detection and instance segmentation.}
Most existing works tackle the problem of ``long-tailed'' distributions in the model training phase, by developing training objectives or algorithms~\cite{wang2020frustratingly,wang2020devil,kang2019decoupling,hu2020learning,ren2020balanced,li2020overcoming,zhang2021distribution,wang2021adaptive}. They usually first pre-train the models in a conventional way, using data from all or just the head classes, and then fine-tunes them on the entire long-tailed dataset using either re-sampling~\cite{gupta2019lvis,chang2021image,shen2016relay} or cost-sensitive learning~\cite{tan2020equalization,tan2020equalizationv2,hsieh2021droploss,wang2020seesaw}. Instead of directly learning a model from long-tailed data, another thread of works investigate data augmentation techniques to improve the performance of long-tailed object detection and instance segmentation~\cite{ghiasi2020simple,ramanathan2020dlwl,zhang2021simple,zang2021fasa}. For example, Simple Copy-Paste~\cite{ghiasi2020simple} augments the training data in the image space using the original long-tailed dataset. FASA~\cite{zang2021fasa} enhances class-wise features using a Gaussian prior. DLWL~\cite{ramanathan2020dlwl} and MosaicOS~\cite{zhang2021simple} extensively leverage extra data sources from YFCC-100M~\cite{thomee2015yfcc100m}, ImageNet~\cite{deng2009imagenet} or Internet to augment the long-tailed LVIS dataset~\cite{gupta2019lvis}.

Our work follows the second thread on learning with additional weakly-supervised or unsupervised data, similar to the recently proposed MosaicOS framework~\cite{zhang2021simple}. We, however, further develop an effective way to obtain high-quality instance segments from object-centric images, while MosaicOS merely learns with pseudo bounding box annotations. Since collecting pixel-level annotations is more challenging and prone to error, we develop a novel ranking mechanism such that only the high-quality segments will be used for model training. Moreover, by copying and pasting the segments into the context of scene-centric images, our method can further bridge the domain gap between different data sources. Overall, we view our approach as a critical leap upon \cite{zhang2021simple,ghiasi2020simple} that can significantly improve long-tailed instance segmentation by largely increasing the training segments of rare objects.

\mypara{Image-based foreground object segmentation.}
There are a variety of techniques that we can potentially leverage to extract the foreground object segments from {object-centric images} without laborious annotations. Representative methods include image (co-)saliency detection~\cite{itti1998model,cheng2014global}, unsupervised/weakly-supervised object segmentation~\cite{rother2004grabcut}, attention~\cite{caron2021emerging}, instance localization~\cite{zhou2016learning,selvaraju2017grad}, and image co-segmentation~\cite{rother2006cosegmentation,chen2020show,zhang2020deep}. The purpose of our work is thus not to propose a new way or compare to those methods, but to investigate approaches that are more effective and efficient for the large-scale long-tailed setting. In this paper, we mainly focus on one potential solution for segmenting foreground objects: image co-segmentation. Aiming at jointly segmenting the common foreground regions from a group of images, co-segmentation is very useful in many semantic labeling tasks~\cite{cech2010efficient,chen2020show} and is a direct fit to the object-centric images we collect. 
Existing image co-segmentation models are usually trained and evaluated on relatively small-scale benchmarks such as MSRC~\cite{shotton2006textonboost}, Internet~\cite{rubinstein2013unsupervised}, iCoseg~\cite{batra2010icoseg}, PASCAL-VOC~\cite{everingham2010pascal}, etc. Our work is almost the first attempt to test the generalizability of existing, pre-trained co-segmentation models on a much larger-scale setting that contains more than $1,000$ categories; each category consists of hundreds or thousands of object-centric images with various appearances and poses. As will be shown in the experimental results and analyses, our framework can effectively utilize the off-the-shelf image co-segmentation models. 

\section{Approach}
\label{s_approach}

\begin{figure*}[t]
    \setlength{\abovecaptionskip}{2mm}
    \setlength{\belowcaptionskip}{-3mm}
    \centerline{\includegraphics[width=1\linewidth]{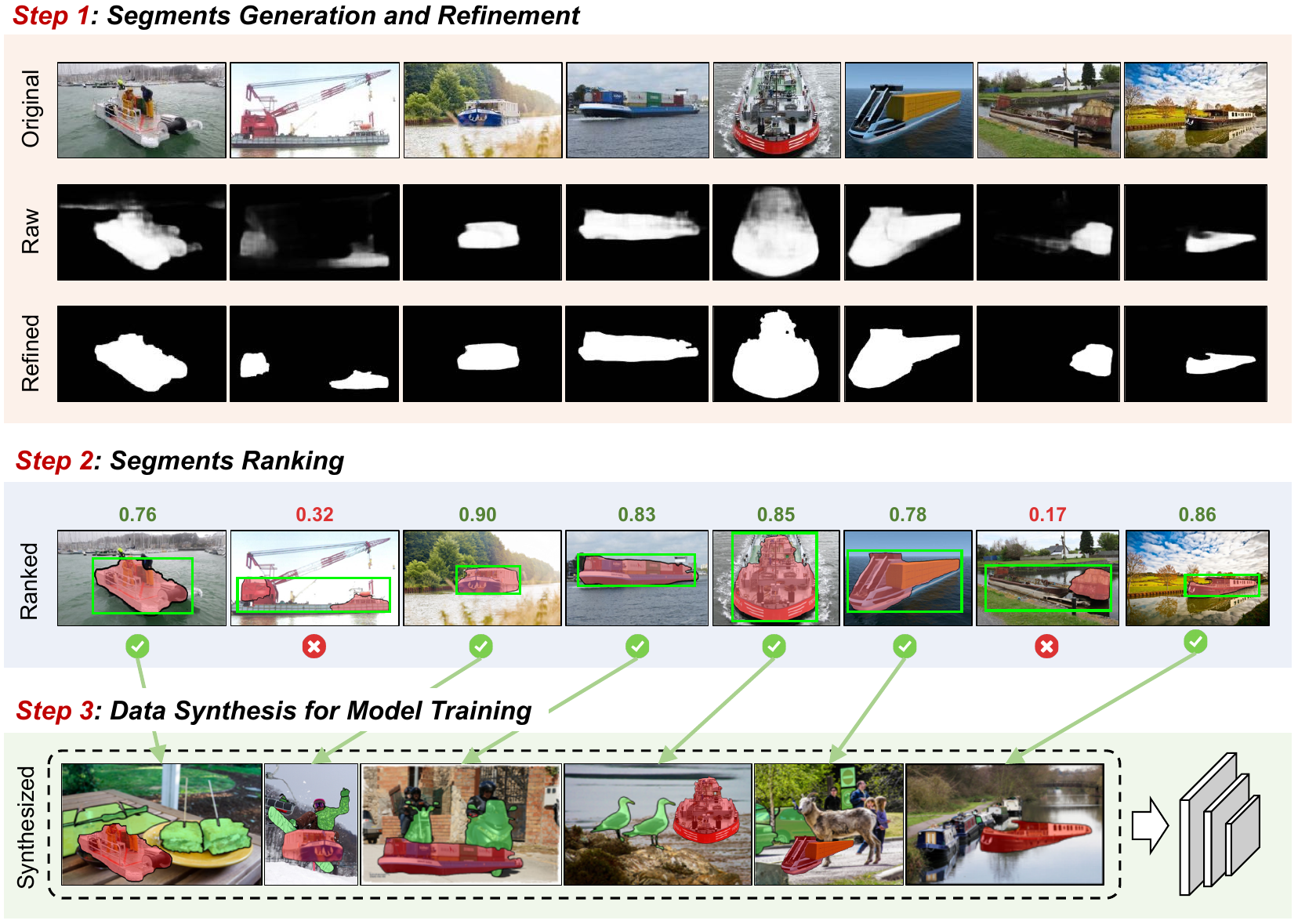}}
    \caption{\small \textbf{Illustration of the \ourmethod pipeline.} We show a rare class \textit{Barge} in LVIS v1~\cite{gupta2019lvis} as the example. We first perform image co-segmentation on top of the object-centric images of \textit{Barge} (outside LVIS v1) to obtain raw object segments, followed by segments refinement. The segments are then scored by a learned ranker (the {\color{green}{green}} boxes in step 2) such that only the high-quality ones would be used for augmenting data for model training. Finally, we randomly paste the selected object segments ({\color{red}{red}}) onto the original scene-centric images of LVIS v1 to improve the long-tailed instance segmentation. {\color{green}{Green}} segments indicate the original objects in scene-centric images.
    }
    \label{fig:pipeline}
\end{figure*}

Our \ourmethod (Free Object Segments) framework for data augmentation is fairly simple and scalable for large-vocabulary and long-tailed instance segmentation.  \autoref{fig:pipeline} illustrates the overall pipeline, which consists of three major steps: (i) segment generation and refinement, (ii) segment ranking,  and (iii) data synthesis for model training.

\subsection{Generating Object Segments}
\label{ss_generation}
We assume that we can obtain a sufficient amount of object-centric images for each class of interest. As discussed in~\cite{zhang2021simple}, this is mostly doable.
We can take advantage of existing image classification datasets like ImageNet~\cite{deng2009imagenet} or leverage image search engine (\eg, Google Images).

\mypara{Raw segments generation.}  
Given object-centric images of the same class, \emph{which usually share similar appearances or contrasts to the background}, we apply image co-segmentation techniques~\cite{rother2006cosegmentation,zhang2020deep} to extract their common foreground regions.
Without loss of generality, we use the state-of-the-art image co-segmentation  algorithms, Spatial and Semantic Modulation (SSM)~\cite{zhang2020deep}.  The outputs of SSM are raw segments in gray scales for each image, as shown in \autoref{fig:pipeline} (see \autoref{ss_setup} for more details). Please be referred to \autoref{s_related} for other potential algorithms for this stage.

\mypara{Post-processing for segment refinement.}
To turn the raw, grayscale segmentation map into a binary one that can be used to train a segmentation model, we threshold the map. 
As the suitable threshold value may vary across images and classes, we apply a Gaussian filter followed by dynamic thresholding, \ie, Li thresholding~\cite{li1993minimum,li1998iterative}, which minimizes the cross-entropy between the foreground and the background to find the optimal threshold to distinguish them. To further improve the resulting binary map, we apply erosion and dilation to smooth the boundary. Finally, we then remove small, likely false positive segments by only keeping the largest connected component in the binary map. \autoref{fig:pipeline} (Step 1) gives an illustration. Please also see supplementary materials for more details.

\subsection{Learning to Rank the Segments}
\label{ss_ranking}

\begin{wrapfigure}{r}{0.53\textwidth}
    \vskip -25pt
    \centering
    \setlength{\abovecaptionskip}{-5mm}
    \setlength{\belowcaptionskip}{-7mm}
    \includegraphics[width=0.53\textwidth]{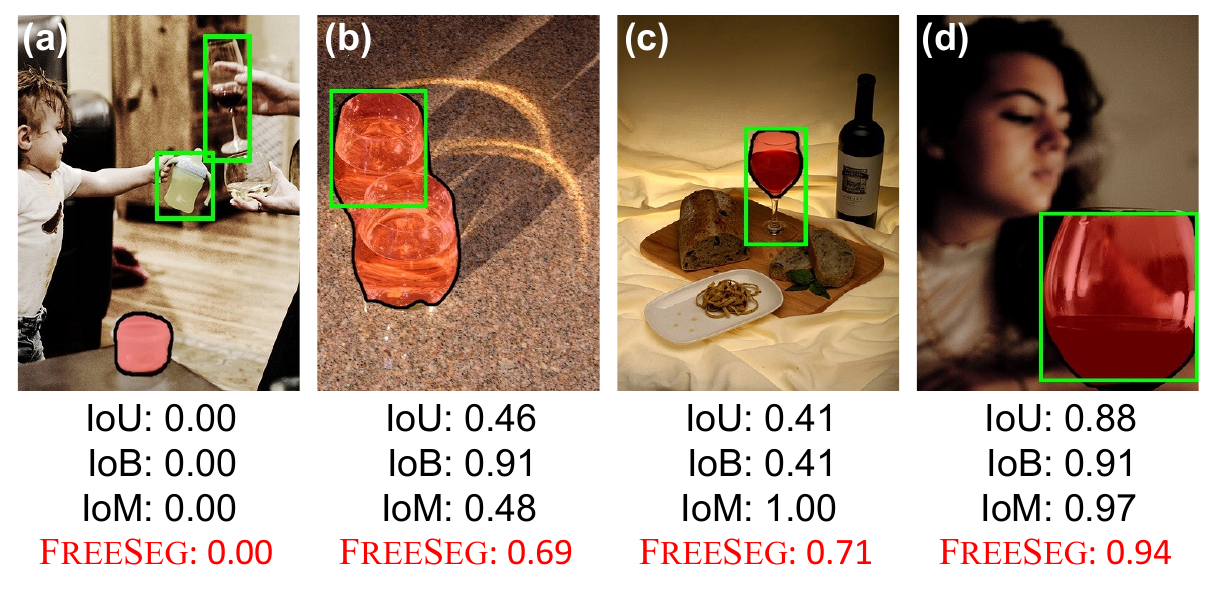}
    \caption{\small \textbf{Comparison of metrics for ranking segments.} We show four examples of the class \textit{wine glasses}. The {\color{red}red} masks are by our method; {\color{green}green} boxes are by LORE.
    In (a) and (d), IoU ranks the segments well, when the box locations are precise. 
    However, in (b), the poor box location leads to a small IoU, even if the segment is precise. In (c), IoU fails due to the specific shape of \textit{wine glasses}, even if the segment is precise. \ourmethod score is able to take all the above into account to faithfully rank segments.
    }
    \label{fig:metrics}
\end{wrapfigure}

While the post-processing step has greatly improved the binary masks and made them look more like the true object masks, they may occasionally miss the target objects (\ie, the objects of the image labels) or include background pixels. This is not surprising: we apply image co-segmentation class-by-class to only explore the within-class similarity. Some co-occurring objects (\eg, persons for unicycles) thus may be miss-identified as the target objects; some target objects that are too small may be dominated by other objects.

At first glance, this seems to paint a grim picture. However, as mentioned in \autoref{s_intro}, our ultimate goal is to \emph{obtain a set of high-quality instance segments from object-centric images, not to segment all the object-centric images well}. Therefore, in the second stage, we develop a novel approach to rank the object segments for each class. Specifically, we aim to select images whose masks truly cover the target objects and are as tight as possible to them.

\mypara{Ranking by learning a classifier.} Given an object segment obtained from co-segmentation, how can we determine if the segment truly covers the target object? Here, we take one intuition: \emph{if a segment covers the target object, then by removing it from the image, an image classifier\footnote{We have image labels for object-centric images, and thus we can train an image classifier upon them.} will unlikely classify the manipulated image correctly.} This idea has indeed been used in~\cite{zhang2021simple} to discover pseudo bounding boxes given only image labels. More specifically, the authors developed ``localization by region removal (LORE)'', which sequentially removes bounding box regions from an image till the image classifier fails to predict the right class. Those removed bounding boxes are then treated as pseudo bounding boxes for the target object class. 

We thus adopt the idea of LORE to rank our object segments. But instead of removing the discovered segments and checking the classifier's failure, we directly compare our object segments to the bounding boxes selected by LORE. \emph{In essence, if the LORE boxes and our segments are highly overlapped, then the segments are considered high-quality.}

\mypara{Ranking metrics.}
Arguably the most common way to characterize the overlap/agreement between two masks/boxes is intersection over union (IoU), which simply treats all contents in a box or mask as foreground. However, this metric is not suitable in our case for the following reasons: (i) both boxes and segments may be noisy, and simply measuring the IoU between them fails to rank good segments when the boxes are poor; 
(ii) object shapes are not always convex, and thus IoU may underestimate the agreement. As shown in \autoref{fig:metrics}, IoU fails to recall true positives. 

\begin{wrapfigure}{r}{0.63\textwidth}
    \vskip -23pt
    \setlength{\abovecaptionskip}{1mm}
    \setlength{\belowcaptionskip}{-8mm}
    \centerline{\includegraphics[width=1\linewidth]{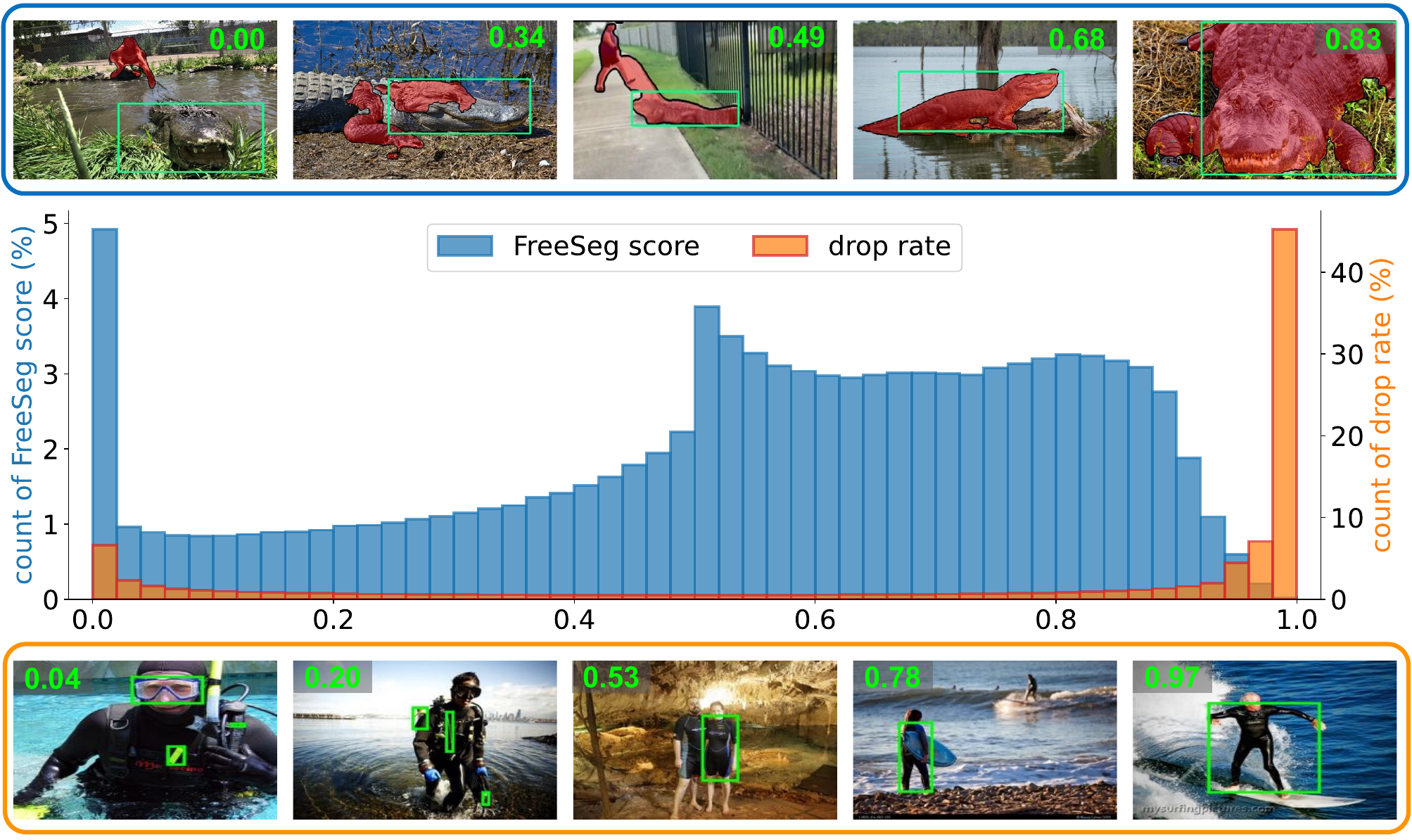}}
    \caption{\small \textbf{Ranking the object segments.} We apply the \ourmethod score and the drop rate to select high-quality segments/images. We show the LORE boxes in green and discovered segments in red. In the top images of the class \textit{alligator}, \ourmethod scores on the upper right corner of images imply the alignment between the segments and the box locations. In the bottom images of the class \textit{wet suit}, drop rates on the upper left corner of images indicate the quality of object-centric images (the larger, the better). These metrics are shown to be effective to rank the segments/images. For both metrics, we simply set a threshold $0.5$ to discard low-confidence segments and images. 
    }
    \label{fig:ranking}
\end{wrapfigure}

We, therefore, propose the \ourmethod score to rank the segments.
We make one mild assumption: either the object box or the segment is trustable, and introduce two metrics: intersection over bounding box (IoB) and intersection over mask (IoM). While they share the same numerator with intersection over union (IoU), they have different denominators. IoM implies that the bounding box is precise and measures how much portion of the mask is inside the box, and vice versa for IoB. We take both into account by averaging them as our \ourmethod score. As shown in \autoref{fig:ranking}, it effectively keeps the good segments in the pool. 

\mypara{Drop rate by the classifier.} We introduce another metric for ranking the segment or, more precisely, its corresponding object-centric image. The rationale is if an object does not clearly show up in an image (\eg, occluded or of small sizes), then the obtained segment is unlikely accurate. To this end, we leverage the image classifier trained for LORE, and compute the drop rate --- the classifier's relative confidence drop for the target class, before and after LORE box removal. Let $s(c)$ and $s'(c)$ denote the classifier's confidence of the target class $c$ before and after LORE box removal from the image, the drop rate is $\frac{s(c)-s'(c)}{s(c)}$.
The drop rate indicates how easily, by removing LORE's localized target objects, would the classifier's confidence reduce.
The larger the drop is, the easier the localization of target objects is, and thus the higher quality the object-centric image is. See \autoref{fig:ranking} for an illustration on the drop rate.

\mypara{Ranking the segments.}
We use both the \ourmethod score and the drop rate to rank the object-centric images and their co-segmentation segments. We keep those with both scores larger than $0.5$ as the high-quality segments.

\subsection{Putting the Segments in the Context}
\begin{figure}[t]
    \setlength{\abovecaptionskip}{2mm}
    \setlength{\belowcaptionskip}{-2mm}
    \centerline{\includegraphics[width=0.99\linewidth]{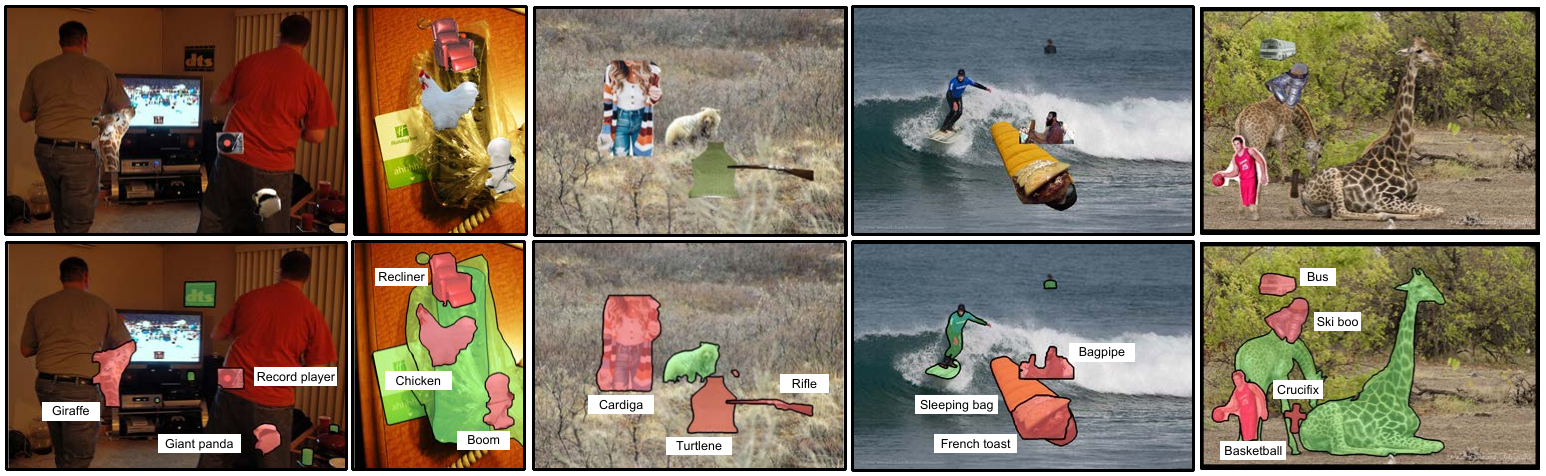}}
    \caption{\small \textbf{Synthesized examples via \ourmethod.} We generate object segments from object-centric images and randomly paste them onto scene-centric images. {\color{red}{Red}} masks indicate pasted segments by \ourmethod; {\color{green}{green}} masks indicate original objects in scene-centric images. Please see supplementary materials for more examples.
    }
    \label{fig:example}
\end{figure}

We now describe how we leverage the discovered high-quality instance segments to facilitate segmentation model training. As discussed in \autoref{s_intro}, instead of directly training the model with the segments on top of object-centric images, we choose to synthesize more scene-centric alike examples by pasting the segments into labeled scene-centric images (\eg, those in LVIS v1~\cite{gupta2019lvis}). We adopt the idea of simple copy-paste augmentation~\cite{ghiasi2020simple} for this purpose. Specifically, we randomly (i) sample several object-centric images, (ii) re-scale and horizontally flip the object segments, and (iii) paste them onto the scene-centric images from the original training set (see \autoref{ss_setup} for details). The resulting synthesized images (see \autoref{fig:example}) then can be used to improve model training.

\section{Experiments}
\label{s_exp}

\subsection{Setup}
\label{ss_setup}

\mypara{Dataset and evaluation metrics.}
We validate our approach on the LVIS v1 instance segmentation benchmark~\cite{gupta2019lvis}. (See the supplementary materials for the results on COCO-LT~\cite{wang2020devil}.) The dataset contains $1,203$ entry-level object categories with around $2$ million high-quality instance annotations. The training set contains $100,170$ images for all the classes; the validation set contains $19,809$ images for $1,035$ classes. The categories follow a long-tailed distribution and are divided into three groups based on the number of training images: rare (1-10 images), common (11-100 images), and frequent ($>$100 images). \emph{We report our results on the validation set by convention.} We adopt the standard mean average precision (AP) metric~\cite{gupta2019lvis}, which sets the cap of detected objects per image to $300$. We denote the AP for rare, common, and frequent classes as $\text{AP}_{r}$, $\text{AP}_{c}$, and $\text{AP}_{f}$, respectively. We also report AP for bounding boxes (\ie, $\text{AP}^{b}$), predicted by the same instance segmentation models. Following~\cite{wu2019detectron2,gupta2019lvis}, we set the score threshold to $1\times10^{-4}$ during testing. No test-time augmentation is used.

\mypara{Object-centric data sources.}
We follow \cite{zhang2021simple} to search images in ImageNet-22K~\cite{deng2009imagenet} and Google Images~\cite{google}. Specifically, we use the unique WordNet synset ID \cite{miller1995wordnet} to match the categories between ImageNet-22K and LVIS v1. We are able to match $997$ LVIS classes and retrieve $1,242,180$ images from ImageNet. Because ImageNet images are nearly balanced by design, with around 1K images/class, the imbalance situation in LVIS can be largely reduced. In addition, we retrieve images via Google by querying with class names provided by LVIS. Such a search returns hundreds of iconic images and we take top $500$ for each of the $1,203$ classes. Overall, for the rarest class (one image in LVIS), the increase factor is larger than 500 times. Please see the supplementary material for more details.

\mypara{Image co-segmentation algorithm.}
We adopt the state-of-the-art image co-segmentation algorithm Spatial and Semantic Modulation (SSM)~\cite{zhang2020deep} to discover raw segments of objects from the object-centric images.
SSM designs a spatial and semantic modulated deep network to jointly learn the structural and semantic information from the  objects in the same class. The checkpoint of the released SSM model  is pre-trained on COCO-SEG dataset~\cite{wang2019robust} with a VGG16 backbone~\cite{simonyan2014very}. We directly apply the model on all the object-centric images for each category without bells and whistles.
 
\mypara{Learning an object segments ranker.}
As mentioned in \autoref{ss_ranking}, we train a $1203$-way classifier with a ResNet-50 backbone~\cite{he2016deep}, using all the object-centric images, to rank the candidate segments within each class. We use a batch size 256 and follow the standard training schedule. The classifier achieves 85\% Top-1 accuracy on the training images. We use the idea of ``localization by region removal'' (LORE)~\cite{zhang2021simple} to detect the bounding boxes of objects. \autoref{tab:stats} shows the statistics of the augmented data before and after the ranking via \ourmethod.

\begin{table}[t]
\small
\tabcolsep 10pt
\centering
\caption{\small \textbf{Statistics of LVIS training data and the augmented data by \ourmethod.} Collected: \# of all images collected from ImageNet-22K and Google Images. Selected via \ourmethod: \# of remaining images selected by segments ranking. Note that our data curation process is quite straightforward and fully automated.}
\label{tab:stats}
\begin{tabular}{rrrr}
\toprule
\textbf{\# of samples} & \textbf{ImageNet} & \textbf{Google Images} & \textbf{Total} \\
\midrule
Original instance  & -- & -- & 1,270K \\
Original image & -- & -- & 100K \\
Collected & 1,242K & 588K & 1,830K \\
Selected via \ourmethod  & 662K & 304K & 966K \\
\bottomrule
\end{tabular}
\vspace{-4mm}
\end{table}

\mypara{Base models for instance segmentation.}
We mainly evaluate the performance of \ourmethod using two base models for instance segmentation, \ie, Mask R-CNN~\cite{he2017mask} and MosaicOS~\cite{zhang2021simple}, implemented with~\cite{wu2019detectron2}.
Both models use ResNet \cite{he2016deep}, which is pre-trained on ImageNet~\cite{russakovsky2015imagenet}, with a Feature Pyramid Network (FPN) \cite{lin2017feature} as the backbone. The base Mask R-CNN model is trained with the LVIS v1 training set with \emph{repeated factor sampling} and follows the standard training procedure in \cite{gupta2019lvis} (1$\times$ scheduler). 

MosaicOS~\cite{zhang2021simple} is one of the state-of-the-art models\footnote{We note that \ourmethod is detector-agnostic and is thus complementary to and compatible with other models~\cite{redmon2017yolo9000,sohn2020simple,zhou2022detecting} that incorporate external images like \cite{zhang2021simple}.}, which is further pre-trained with balanced object-centric images from ImageNet-22K and Google Images. However, MosaicOS mainly focuses on improving long-tailed object detection with pseudo-labeled bounding boxes. As will be shown in the experimental results, \ourmethod can notably boost the performance upon MosaicOS with the same image resources. Furthermore, such an improvement can not be achieved by the vanilla simple copy-paste~\cite{ghiasi2020simple} using the training data from LVIS alone, especially for rare object categories.

\mypara{Details of object segments pasting.}
We follow the pasting mechanism in \cite{ghiasi2020simple} to randomly pick examples from LVIS training set as the background images. We then paste segments from $N$ random object-centric images at different locations of each background image, where $N$ is in [1, 6]\footnote{The median number of instances per image in LVIS dataset is 6.}. For LVIS images, we follow the standard data augmentation policy in \cite{gupta2019lvis} and \cite{wu2019detectron2}. For binary masks originally on LVIS images, we remove pixel annotations if the objects are occluded by the pasted ones in the front. Please see the supplementary material for more details.

\mypara{Training and optimization.}
Given the base instance segmentation model, we first fine-tune the model for 90K iterations with \ourmethod segments, using all the loss terms in Mask R-CNN. We fine-tune all the parameters except the batch-norm layers in the backbone. 
We then fine-tune the model again for another 90K iterations using the original LVIS training images. The rationale of training with multiple stages is to prevent the augmented instances from dominating the training process (see \autoref{tab:stats} for statistics) and it is shown to be effective in \cite{zhang2021simple}. Both fine-tuning steps are trained with stochastic gradient descent with a batch size of 8, momentum of 0.9, weight decay of $10^{-4}$, and learning rate of $2\times10^{-4}$. All models are trained with four NVIDIA A6000 GPUs.

\begin{table}[t]
\small
\tabcolsep 7pt
\centering
\caption{\small \textbf{State-of-the-art comparison on LVIS v1 instance segmentation.} \ourmethod are initialized with MosaicOS~\cite{zhang2021simple} as the base model. ${2\times}$: Seesaw applies a stronger ${2\times}$ training schedule while other methods are with ${1\times}$ schedule. $\star$: with post-processing calibration introduced by \cite{pan2021model}.}
\label{tab:sota}
\begin{tabular}{clccccc}
\toprule
\textbf{Backbone} & \multicolumn{1}{c}{\textbf{Method}} & \textbf{AP} & \textbf{$\text{AP}_{r}$} & \textbf{$\text{AP}_{c}$} & \textbf{$\text{AP}_{f}$} & \textbf{$\text{AP}^{b}$} \\
\midrule
\multirow{12}{1.9cm}{\centering ResNet-50 FPN} & RFS~\cite{gupta2019lvis} & 22.58 & 12.30 & 21.28 & 28.55 & 23.25 \\
& BaGS~\cite{li2020overcoming} & 23.10 & 13.10 & 22.50 & 28.20 & 25.76 \\
& Forest R-CNN~\cite{wu2020forest} & 23.20 & 14.20 & 22.70 & 27.70 & 24.60 \\
& RIO~\cite{chang2021image} & 23.70 & 15.20 & 22.50 & 28.80 & 24.10 \\
& EQL v2~\cite{tan2020equalizationv2} & 23.70 & 14.90 & 22.80 & 28.60 & 24.20 \\
& FASA~\cite{zang2021fasa} & 24.10 & 17.30 & 22.90 & 28.50 & -- \\
& DisAlign~\cite{zhang2021distribution} & 24.30 & 8.50 & 26.30 & 28.10 & 23.90 \\
& Seesaw~\cite{wang2020seesaw}$^{2\times}$ & 26.40 & 19.60 & 26.10 & 29.80 & 27.40 \\
\cdashline{2-7}
& MosaicOS~\cite{zhang2021simple} & 24.45 & 18.17 & 23.00 & 28.83 & 25.05 \\
& \cellcolor{LightCyan}{w/ \ourmethod} & \cellcolor{LightCyan}{25.19} & \cellcolor{LightCyan}{\textit{\textbf{20.23}}} & \cellcolor{LightCyan}{23.80} & \cellcolor{LightCyan}{28.92}  & \cellcolor{LightCyan}{25.98}  \\
\cdashline{2-7}
& MosaicOS~\cite{zhang2021simple}~$\star$ & 26.76 & 23.86 & 25.82 & 29.10 & 27.77 \\
& \cellcolor{LightCyan}{w/ \ourmethod$\star$} & \cellcolor{LightCyan}{27.34} & \cellcolor{LightCyan}{\textit{\textbf{25.11}}} & \cellcolor{LightCyan}{26.29} & \cellcolor{LightCyan}{29.49} & \cellcolor{LightCyan}{28.47}\\
\midrule
\multirow{7}{1.9cm}{\centering ResNet-101 FPN} & RFS~\cite{gupta2019lvis} & 24.82 & 15.18 & 23.71 & 30.31 & 25.45 \\
& FASA~\cite{zang2021fasa} & 26.30 & 19.10 & 25.40 & 30.60 & -- \\
& Seesaw~\cite{wang2020seesaw}$^{2\times}$ & 28.10 & 20.00 & 28.00 & 31.90 & 28.90\\
\cdashline{2-7}
& MosaicOS~\cite{zhang2021simple} & 26.73 & 20.52 & 25.78 & 30.53 & 27.41 \\
& \cellcolor{LightCyan}{w/ \ourmethod} & \cellcolor{LightCyan}{27.54} & \cellcolor{LightCyan}{\textit{\textbf{23.00}}}  & \cellcolor{LightCyan}{26.48}  & \cellcolor{LightCyan}{30.72}   & \cellcolor{LightCyan}{28.63}  \\
\cdashline{2-7}
& MosaicOS~\cite{zhang2021simple}~$\star$ & 29.03 & 26.38 & 28.15 & 31.19 & 29.96 \\
& \cellcolor{LightCyan}{w/ \ourmethod$\star$} & \cellcolor{LightCyan}{29.72} &  \cellcolor{LightCyan}{\textit{\textbf{28.69}}} &  \cellcolor{LightCyan}{28.67} &  \cellcolor{LightCyan}{31.34} & \cellcolor{LightCyan}{31.11} \\
\midrule
\multirow{5}{1.9cm}{\centering ResNeXt-101 FPN} & RFS~\cite{gupta2019lvis} & 26.67 & 17.60 & 25.58 & 31.89 & 27.35  \\
\cdashline{2-7}
& MosaicOS~\cite{zhang2021simple} & 28.29 & 21.75 & 27.22 & 32.35 & 28.85 \\
& \cellcolor{LightCyan}{w/ \ourmethod} & \cellcolor{LightCyan}{28.86} &  \cellcolor{LightCyan}{\textit{\textbf{23.34}}} &  \cellcolor{LightCyan}{27.77} &  \cellcolor{LightCyan}{32.49} & \cellcolor{LightCyan}{29.98} \\
\cdashline{2-7}
& MosaicOS~\cite{zhang2021simple}~$\star$ & 29.81 & 25.73 & 28.92 & 32.59 & 30.56  \\
& \cellcolor{LightCyan}{w/ \ourmethod$\star$} & \cellcolor{LightCyan}{30.37} &  \cellcolor{LightCyan}{\textit{\textbf{26.43}}} &  \cellcolor{LightCyan}{29.63} &  \cellcolor{LightCyan}{32.92} & \cellcolor{LightCyan}{31.81} \\
\bottomrule
\end{tabular}
\end{table}

\subsection{Main Results on Instance Segmentation}

\mypara{State-of-the-art comparison.}
We compare to the state-of-the-art methods for long-tailed instance segmentation in \autoref{tab:sota}. The proposed \ourmethod method achieves comparable or even better results, especially for rare object categories. For example, \ourmethod outperforms all the other methods except Seesaw loss~\cite{wang2020seesaw}, which is implemented with a different framework~\cite{mmdetection} and trained with a stronger scheduler. (We provide further comparisons in this aspect in the supplementary.)

\mypara{Backbone agnostic.}
Beyond ResNet-50, we further evaluate \ourmethod with stronger backbone model architectures: ResNet-101~\cite{he2016deep} and ResNeXt-101~\cite{xie2017aggregated}, following the same training pipeline as ResNet-50. \ourmethod achieves notable gains over MosaicOS~\cite{zhang2021simple}, justifying that \ourmethod can benefit different instance segmentation models and architectures.

\mypara{Compatibility with existing methods.}
We further apply post-processing calibration~\cite{pan2021model} on top of the model trained with \ourmethod. Results are shown in \autoref{tab:sota} (\ourmethod$\star$) and the improvements are consistent. More surprisingly, \ourmethod can boost the performance of rare classes to be similar to common classes. This indicates that by introducing more while not so perfect training instances, \ourmethod dramatically overcomes the long-tailed problem.

\subsection{Detailed Analyses and Ablation Studies}

\mypara{Does segment ranking help?} 
The quality of the segments is important because inferior pixel-level annotations for instance segmentation may contain certain noise (cf. \autoref{s_intro}). Such an issue will be amplified for rare categories when the training examples are long-tailed. 
Here we conduct experiments with and without ranking object segments. As shown in \autoref{tab:stats}, we are able to collect $1,830$K segments from ImageNet-22k and Google Images, while only half of them are left after filtering with \ourmethod. \autoref{tab:ablation} shows the results. While both versions outperform the baseline models, segment ranking does help more (row 4 \emph{vs.} row 2 in \autoref{tab:ablation}), suggesting that the quality of pixel labels is more important than the quantity for instance segmentation.

We notice that filtering by ranking gives higher quality but fewer masks. To further understand the effect of quality and quantity of object segments on the accuracy of \ourmethod, we randomly sample the original co-segmentation masks such that the remaining ones are of the same quantity as those \emph{selected by our ranking method}. We see a bigger gain by our ranking method (row 4 \emph{vs.} row 3 in \autoref{tab:ablation}), justifying its effectiveness in selecting high-quality masks.

\begin{table}[t]
\small
\tabcolsep 4pt
\centering
\caption{\small \textbf{Ablation study on object segments ranking.} We evaluate the performance of the model trained with and without the segments ranking mechanism by \ourmethod. Results demonstrate the importance of ranking the object segments.}
\label{tab:ablation}
\begin{tabular}{lccrcccc}
\toprule
\multicolumn{1}{c}{\textbf{Method}} & \multicolumn{1}{c}{\textbf{Random}} & \multicolumn{1}{c}{\textbf{Ranking}}  & \textbf{\#Image} & \textbf{AP} & \textbf{$\text{AP}_{r}$} & \textbf{$\text{AP}_{c}$} & \textbf{$\text{AP}_{f}$}  \\
\midrule
MosaicOS~\cite{zhang2021simple} &  &  & & {24.45} & {18.17} & {23.00} & {28.83} \\
\multirow{3}{*}{w/ \ourmethod} &  & & 1,830K& 24.87 & 19.13 & 23.55  & 28.86   \\
& \cmark & & 966K & 24.50 & 18.68 & 23.18  & 28.52   \\
& &\cmark & 966K & \textbf{25.19} & \textbf{20.23} & \textbf{23.80} & \textbf{28.92} \\
\bottomrule
\end{tabular}
\end{table}

\begin{table}[t]
\small
\tabcolsep 8pt
\centering
\caption{\small \textbf{Analysis on different object segments ranking metrics.} The proposed \ourmethod score can take different scenarios into account thus achieving better results.}
\label{tab:metrics}
\begin{tabular}{lccccc}
\toprule
\multicolumn{1}{c}{\textbf{Method}} & \textbf{Ranking Metrics} & \textbf{AP} & \textbf{$\text{AP}_{r}$} & \textbf{$\text{AP}_{c}$} & \textbf{$\text{AP}_{f}$}  \\
\midrule
MosaicOS~\cite{zhang2021simple} & -- &  24.45 & 18.17 & 23.00 & 28.83 \\
\multirow{4}{*}{w/ \ourmethod} & IoU & 24.74  & 19.04 & 23.58  & 28.53  \\
&  IoB & 24.69  & 18.41	& 23.58 & 28.70 \\
&  IoM & 24.56  & 18.62	& 23.14 & 28.74  \\
& \ourmethod & \textbf{25.19} & \textbf{20.23} & \textbf{23.80} & \textbf{28.92}  \\
\bottomrule
\end{tabular}
\end{table}

\mypara{Ranking metrics.} We show both quantitative and qualitative comparisons of different ranking metrics for filtering noisy segments in \autoref{tab:metrics} and \autoref{fig:metrics}. \ourmethod score can take different scenarios into account and successfully select confident segments from noisy ones. This verifies that the quality of the segments is the key and that the proposed \ourmethod pipeline effectively does the job.

\mypara{Effect of segments filtering by drop rate.}
\autoref{tab:drop_rate} reports results with and without segments filtering with drop rate (cf. \autoref{ss_ranking}). By jointly using drop rate and \ourmethod score, our method achieves better results by using fewer and cleaner object segments for training.

\begin{table}[t]
\small
\tabcolsep 6pt
\centering
\caption{\small \textbf{Ablation study on segments filtering by drop rate.} }
\label{tab:drop_rate}
\begin{tabular}{lcrcccc}
\toprule
\multicolumn{1}{c}{\textbf{Method}} & \textbf{Drop Rate} & \textbf{\#Image} & \textbf{AP} & \textbf{$\text{AP}_{r}$} & \textbf{$\text{AP}_{c}$} & \textbf{$\text{AP}_{f}$}  \\
\midrule
MosaicOS~\cite{zhang2021simple} &  &  & 24.45 & 18.17 & 23.00 & 28.83 \\
\multirow{2}{*}{w/ \ourmethod} & \xmark & 1,134K & 24.81 & 19.37  & 23.62   & 28.54 \\
& \cmark & 966K & \textbf{25.19} & \textbf{20.23} & \textbf{23.80} & \textbf{28.92}  \\
\bottomrule
\end{tabular}
\end{table}

\begin{table}[t]
\small
\tabcolsep 7pt
\centering
\caption{\small \textbf{Importance of the context.} Segments Pasting~\cite{ghiasi2020simple}: \xmark~indicates directly training the instance segmentation model on the object-centric images, using \ourmethod segments as supervision.}
\label{tab:train_oci}
\begin{tabular}{lccccc}
\toprule
\multicolumn{1}{c}{\textbf{Method}} & \textbf{Segments Pasting} &\textbf{AP} & \textbf{$\text{AP}_{r}$} & \textbf{$\text{AP}_{c}$} & \textbf{$\text{AP}_{f}$}  \\
\midrule
MosaicOS~\cite{zhang2021simple} & & 24.45 & 18.17 & 23.00 & 28.83 \\
\multirow{2}{*}{w/ \ourmethod} & \xmark & 24.78  & 18.85 & 23.40  & 28.90 \\
 & \cmark & \textbf{25.19} & \textbf{20.23} & \textbf{23.80} & \textbf{28.92} \\
\bottomrule
\end{tabular}
\end{table}

\mypara{Importance of the context.}
We investigate training the instance segmentation model directly with the object-centric images without pasting \ourmethod segments to LVIS images. The results are shown in \autoref{tab:train_oci}. As expected, we see that the performance is worse than the proposed \ourmethod framework, in which we apply copy-paste augmentation to put the object instances into the context of original training images. This demonstrates the fact that there exists a gap between the contexts of two different image resources, which could limit the improvement on the main task. 

\mypara{Additional results.} Please see supplementary materials, including the results with other evaluation metrics and datasets, the analysis of multi-stage training, effects of different data sources, \textbf{qualitative results}, etc.

\subsection{Comparison to Pasting Ground-truth Segments}
Ghiasi \etal ~\cite{ghiasi2020simple} show that copying and pasting human-annotated segments from one image to another as augmentation can improve instance segmentation with richer \textit{context diversity}. They employ a much larger batch size and longer scheduler with another strong augmentation, large-scale jittering~\cite{ghiasi2020simple}. However, in this work, we focus on enriching \textit{appearance diversity} for objects with abundant free segments from object-centric images. We, therefore, conduct a detailed comparison between pasting ground truth and \ourmethod object segments. We follow the pasting mechanism in \autoref{ss_setup} but use ground truths segments instead. That is, we randomly pick two images from the LVIS training set, apply the same data augmentation policy following the standard instance segmentation model (\ie, resizing shortest edge and random horizontal flip), and then \emph{paste random numbers of instances from one image onto the other image}. 

We show results in \autoref{tab:main}. We validate \ourmethod on two base models and compare to the results of using ground truth segments for augmentation.
\ourmethod achieves consistent gains against the baseline models and, more importantly, outperforms those with copy-paste augmentation using only ground truth segments. This demonstrates that with ample images that can be acquired easily online, even noisy labels without any human efforts could significantly improve long-tailed instance segmentation. We also note that \ourmethod is more effective when the baseline is already re-balanced (\eg, MosaicOS in \autoref{tab:main} bottom and \autoref{tab:sota}), while GT-only can hardly improve upon it due to the lack of training examples. Furthermore, by learning with copy-paste from both sources, the gain can be even larger on both base models. These observations demonstrate that, besides context diversity, the appearance diversity of objects is also the key to improve segmentation.

\begin{table}[t]
\small
\tabcolsep 6pt
\centering
\caption{\small \textbf{Comparison of pasting ground truth (GT) object segments and \ourmethod.} The base models are trained with ResNet-50 and FPN. $\dagger$: models from \cite{zhang2021simple}.} 
\label{tab:main}
\begin{tabular}{cccccccc}
\toprule
\textbf{Method} & \textbf{GT} & \ourmethod &  \textbf{AP} & \textbf{$\text{AP}_{r}$} & \textbf{$\text{AP}_{c}$} & \textbf{$\text{AP}_{f}$} & \textbf{$\text{AP}^{b}$} \\
\midrule
\multirow{4}{*}{Mask R-CNN~\cite{gupta2019lvis}$\dagger$} &  &  & {22.58} & {12.30} & {21.28} & {28.55} & {23.25} \\
& \cmark &  & 24.06 & 17.00  &22.62 & 28.77&24.91 \\
&  & \cmark & \cellcolor{LightCyan}{24.28}  &  \cellcolor{LightCyan}{{{17.68}}} & \cellcolor{LightCyan}{22.79} & \cellcolor{LightCyan}{28.83} & \cellcolor{LightCyan}{25.13} \\
& \cmark & \cmark & \cellcolor{LightCyan}{24.74}  &  \cellcolor{LightCyan}{\textit{\textbf{18.80}}} & \cellcolor{LightCyan}{23.38} & \cellcolor{LightCyan}{28.86} & \cellcolor{LightCyan}{25.51} \\
\midrule
\multirow{4}{*}{MosaicOS~\cite{zhang2021simple}$\dagger$} &  &  & {24.45} & {18.17} & {23.00} & {28.83} & {25.05} \\
& \cmark &  & 24.57 & 18.63 & 23.31 & 28.59 & 25.52 \\
&  & \cmark & \cellcolor{LightCyan}{25.19} & \cellcolor{LightCyan}{{{20.23}}}	& \cellcolor{LightCyan}{23.80} & \cellcolor{LightCyan}{28.92}  &\cellcolor{LightCyan}{25.98} \\
& \cmark & \cmark & \cellcolor{LightCyan}{25.36}  &  \cellcolor{LightCyan}{\textit{\textbf{20.72}}} & \cellcolor{LightCyan}{24.00} & \cellcolor{LightCyan}{28.92} & \cellcolor{LightCyan}{26.00} \\
\bottomrule
\end{tabular}
\end{table}
\section{Conclusion}
\label{s_disc}

Our main contribution and novelty are the insight that object segments emerge freely from object-centric images, and they effectively benefit the challenging long-tailed instance segmentation problem.
We propose a scalable framework \ourmethod to realize this idea. 
We show that, with the underlying properties of object-centric images, simple co-segmentation with proper ranking can result in high-quality instance segments to largely increase the labeled training instances.

We believe that the prospect of leveraging ample data without human labeling has enormous future potential. We note that there are several ways to realize this insight, and \cite{zhang2020deep} is just an instantiation but turns out to be very useful: it is worth mentioning that co-segmentation has never been used to enhance instance segmentation. Further, our pipeline is clean and conceptually simple, clearly indicating
where future improvement can be made (\eg, segment discovery,
extraction, leveraging). We expect our approach to serving as a strong baseline for this direction: for future work to build upon and take advantage of.

\mypara{Acknowledgments}
This research is supported in part by grants from the National Science Foundation (IIS-2107077, OAC-2118240, OAC-2112606), the OSU CCTS pilot grant, and Cisco Systems, Inc. We are thankful for the generous support of the computational resources by the Ohio Supercomputer Center and AWS Cloud Credits for Research.

\clearpage
%
%
\bibliographystyle{splncs04}
\bibliography{main}

\clearpage
\appendix
\begin{center}
\textbf{\large Supplementary Materials}
\end{center}

\appendix

In this supplementary material, we provide details and results omitted in the main text.

\begin{itemize}
    \item \autoref{suppl_details}: implementation details.
    \item \autoref{supp_new_metric}: results on other metrics: AP fixed and boundary IoU.
    \item \autoref{supp_coco_lt}: results on COCO-LT dataset.
    \item \autoref{suppl_ablation}: additional ablation studies.
    \item \autoref{supp_qual}: qualitative results.
\end{itemize}

\section{Implementation Details}
\label{suppl_details}

\subsection{Data Curation}
As mentioned in Section 4.1 of the main paper, we followed~\cite{zhang2021simple} to collect Google images. We use class names as the keywords and take top images from the respective search engine without extra user interventions. Thus, the curation process is quite straightforward. The collected Google data are balanced (500/class). ImageNet images are nearly balanced by design, with around 1K images/class, including rare objects in LVIS. The imbalance situation in LVIS is largely reduced. For the rarest class (one image in LVIS), the increase factor is larger than 500 times.

\subsection{Generating Object Segments}
As mentioned in Section 3.1 and Section 4.1 of the main paper, we apply spatial and semantic modulation (SSM) co-segmentation method~\cite{zhang2020deep} to the object-centric images for each class, followed by segment refinement. We show more examples of object segments by \ourmethod in \autoref{fig:supp_segment1}, \autoref{fig:supp_segment2}, and \autoref{fig:supp_segment3}. With the proper ranking algorithm, our approach can identify the most reliable instance segments to improve long-tailed instance segmentation.

\begin{SCfigure}[]
  {\includegraphics[width=0.6\linewidth]{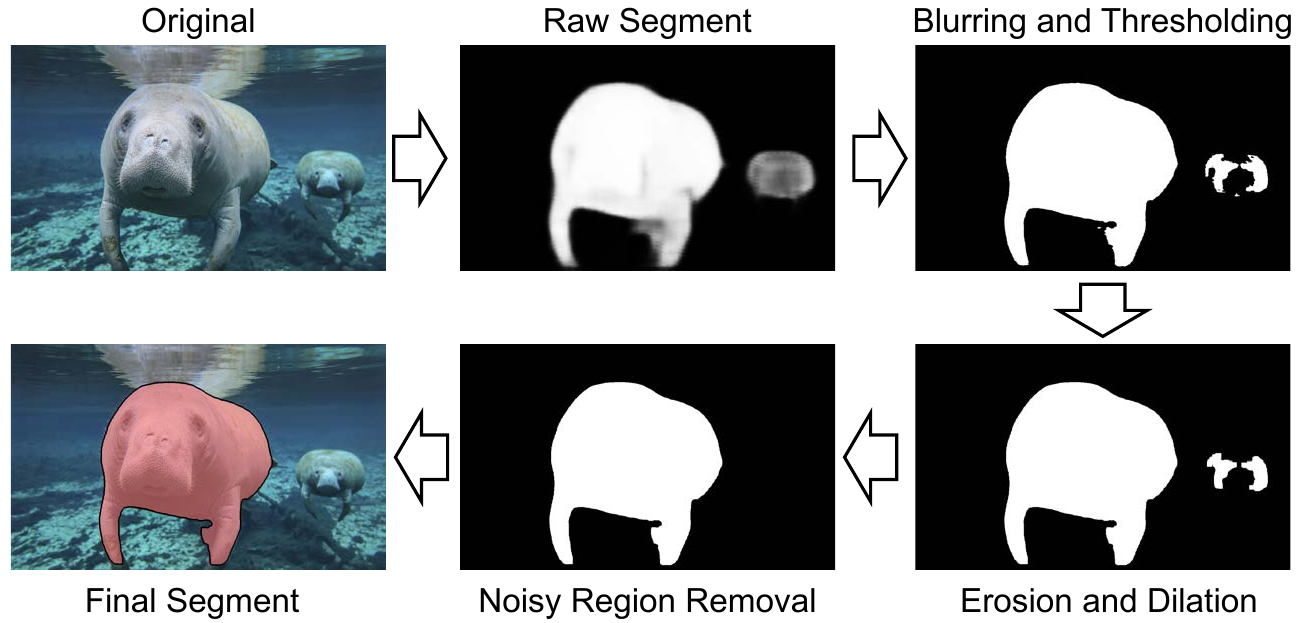}
  \vspace{-3mm}}  
  \caption{\small \textbf{Procedure of generating object segments.} We generate the final segments (\eg, \emph{manatee}) by post-processing the raw segments obtained from the image co-segmentation methods.}
  \label{fig:refinement}
\end{SCfigure}

\subsection{Post-Processing for Segment Refinement}
To turn the raw, gray scale segmentation map into a binary one that can be used to train a segmentation model, we threshold the map. 
As the suitable threshold value may vary across images and classes, we apply Gaussian filter followed by dynamic thresholding, \ie, Li thresholding~\cite{li1993minimum,li1998iterative}, which minimizes the cross-entropy between the foreground and the background to find the optimal threshold to distinguish them.

To further improve the resulting binary map, we apply erosion and dilation to smooth the boundary.  We then remove small, likely false positive segments by only keeping the largest connected component in the binary map. 
\autoref{fig:refinement} shows the entire post-processing procedure for refinement, which greatly improves the quality of the segmentation masks, as illustrated in Figure 2 of the main paper.

\subsection{Putting Segments in Context}
\label{suppl_ss_copy_paste}

As introduced in Section 3.3 and Section 4.1 of the main paper, we follow the mechanism in \cite{ghiasi2020simple} to paste our ranked segments. More specifically, we randomly pick an example from LVIS training set as a background image, followed by pasting segments from 1 to 6 object-centric images on it at different locations. 
For LVIS images, we follow the standard data augmentation policy in \cite{gupta2019lvis} and \cite{wu2019detectron2}. 
That is, we randomly resize the shortest edge of the image into [640, 672, 704, 736, 768, 800] with a limit of max size of width or height to 1333, followed by a random horizontal flip with $p=0.5$. For the selected object-centric images, we apply random horizontal flip ($p=0.5$) followed by random resize with a scale of [0.1, 2.0]. 
We then randomly crop (or pad) the object-centric images to match the size of the background image. Note that, this step ensures that the object segments will be randomly pasted at different locations on each of the LVIS images. 
For binary masks on LVIS images
used for supervision,
we remove pixel annotations if the objects are occluded by the pasted ones in the front. 

The examples of synthesized data via vanilla copy-paste (\ie pasting ground truths) can be found in \autoref{fig:supp_copy_paste}. We also provide examples generated by \ourmethod framework in \autoref{fig:supp_copy_paste_freeseg}. We can see that \ourmethod can increase the appearance diversity of foreground instances, especially for rare object categories. We will leave a better way to leverage the object segments as our future work.

\subsection{Model Training}
We apply a two-stage strategy to fine-tune the pre-trained instance segmentation model (cf. Section 4.1 of the main paper). Both stages follow the same training and optimization setting, which is summarized in \autoref{tab:supp_train_setting}.

\begin{table*}[th]
\small
\tabcolsep 2pt
\centering
\caption{\small \textbf{Optimization configuration for the two-stage fine-tuning.}}
\label{tab:supp_train_setting}
\begin{tabular}{lll}
\toprule
\textbf{Config} & \multicolumn{2}{l}{\textbf{Value}}\\
\midrule
Optimizer & SGD \\
Learning rate & 2e-4 \\
Weight decay & 0.0001 \\
Optimizer momentum & 0.9\\
Batch size & \multicolumn{2}{p{7cm}}{8 (larger batch size, \eg, 16, does not lead to notable differences)} \\
Warm up epoch & 0 \\
Training iteration & 90,000 \\
Aug. for background image & \multicolumn{2}{p{7cm}}{ResizeShortestEdge [640, 672, 704, 736, 768, 800], RandomFlip}\\
Aug. for pasted image &  RandomFlip, ResizeScale [0.1, 2], FixedSizeCrop\\
\bottomrule
\end{tabular}
\end{table*}

\section{Results on AP Fixed with Boundary IoU}
\label{supp_new_metric}

\mypara{\ourmethodbf is effective in AP Fixed with Boundary IoU.}
Besides standard Mask AP, we also report the results in AP Fixed~\cite{dave2021evaluating} with Boundary IoU~\cite{cheng2021boundary}, following the official evaluation metrics in LVIS challenge 2021. AP Fixed replaces the cap (\ie, 300) of number of detected objects per image by a cap (\ie, 10,000) per class for the entire validation set. \autoref{tab:supp_fixed} reports the results. We see that the improvement is consistent, demonstrating \ourmethod is metric-agnostic.

\begin{table}[h]
\small
\tabcolsep 10pt
\centering
\caption{\small \textbf{Results on AP Fixed~\cite{dave2021evaluating} with Boundary IoU~\cite{cheng2021boundary}.} All models are based on ResNet-50 FPN backbone architecture.}
\label{tab:supp_fixed}
\begin{tabular}{lcccc}
\toprule
\multicolumn{1}{c}{\textbf{Method}} & \textbf{AP} & \textbf{$\text{AP}_{r}$} & \textbf{$\text{AP}_{c}$} & \textbf{$\text{AP}_{f}$} \\
\midrule
Mask R-CNN~\cite{gupta2019lvis} & 19.88 & 14.76 & 19.32 & 22.76\\
w/ \ourmethod & \textbf{21.25} & \textbf{18.33} & \textbf{20.85} & \textbf{23.00}  \\
\midrule
MosaicOS~\cite{zhang2021simple} & 21.20 & 18.79 & 20.63 & 22.90 \\
w/ \ourmethod & \textbf{21.86} & \textbf{20.12} & \textbf{21.50} & \textbf{23.03} \\
\bottomrule
\end{tabular}
\end{table}

\section{Results on COCO-LT dataset}
\label{supp_coco_lt}

To further validate the generalizability of our framework, we conduct experiments on another popular long-tailed dataset, \ie, COCO-LT~\cite{wang2020devil}. We match class names to find object-centric images from ImageNet-22K and Google for each class in COCO-LT. We follow the same evaluation protocol in \cite{wang2020devil,zang2021fasa} and show results in \autoref{tab:supp_coco_lt}. \ourmethod (with Mask R-CNN as baseline) outperforms SimCal~\cite{wang2020devil} and FASA~\cite{zang2021fasa}, justifying the generalizability.

\begin{table}[h]
\small
\tabcolsep 6pt
\centering
\caption{\small \textbf{Results on COCO-LT dataset.}}
\label{tab:supp_coco_lt}
\begin{tabular}{lccccc}
\toprule
\multicolumn{1}{c}{\textbf{Method}} &  \textbf{AP} & \textbf{$\text{AP}_{1}$} & \textbf{$\text{AP}_{2}$} & \textbf{$\text{AP}_{3}$} & \textbf{$\text{AP}_{4}$}  \\
\midrule
Mask R-CNN~\cite{zhang2021simple} &  18.70 & 0.00 & 8.20 & 24.40 & 26.00  \\
\midrule
SimCal~\cite{wang2020devil} & 21.80 & 15.00 & 16.20 &  24.30 &  26.00 \\
FASA~\cite{zang2021fasa} & 23.40 & 13.50 & 19.00 & 25.20 & 27.50 \\
\ourmethod & \textbf{25.10} & \textbf{15.80} & \textbf{20.60} & \textbf{27.60} & \textbf{28.80}   \\
\bottomrule
\end{tabular}
\end{table}

\section{Additional Ablation Studies}
\label{suppl_ablation}

\mypara{Effect of data sources}
We first study the effect of data sources. As ImageNet only covers 997 classes of LVIS, we augment it with Google images for all the 1,203 LVIS classes (Section 4.1 of the main paper). \autoref{tab:supp_data} shows results with different data sources, we compare the performance of using different data sources. We see that both Google images and ImageNet are useful. We achieve the best result by combing them.

\begin{table}[h]
\small
\tabcolsep 3pt
\centering
\caption{\small \textbf{Results on different object-centric image sources.} G: Google Images. IN: ImageNet.}
\label{tab:supp_data}
\begin{tabular}{lcccccc}
\toprule
\multicolumn{1}{c}{\textbf{Method}} & \textbf{G} & \textbf{IN} & \textbf{AP} & \textbf{$\text{AP}_{r}$} & \textbf{$\text{AP}_{c}$} & \textbf{$\text{AP}_{f}$}  \\
\midrule
Mask R-CNN~\cite{gupta2019lvis} & & &  22.58 & 12.30 & 21.28 & 28.55 \\
\midrule
\multirow{3}{*}{w/ \ourmethod} & \cmark & & 24.08  & 17.08 & 22.68  & 28.72  \\
&  & \cmark & 24.12  & 17.23 & 22.67  & 28.75  \\
& \cmark & \cmark & \textbf{24.28} & \textbf{17.68} & \textbf{22.79} & \textbf{28.83}  \\
\bottomrule
\end{tabular}
\end{table}

\mypara{Importance of multi-stage training}
\autoref{tab:supp_multi_stage} reports results after the first and second stage training (cf. Section 4.1 of the main paper). As introduced in~\cite{zhang2021simple}, the first stage learns better features with diverse and balanced data, but noisy labels; the second stage trained with accurate labels helps correct the prediction. We note that both stages use repeat factor sampling~\cite{gupta2019lvis} to further balance data.

\begin{table}[h]
\small
\tabcolsep 4pt
\centering
\caption{\small \textbf{Importance of multi-stage training.}}
\label{tab:supp_multi_stage}
\begin{tabular}{lccccc}
\toprule
\multicolumn{1}{c}{\textbf{Method}} & \textbf{Stage} & \textbf{AP} & \textbf{$\text{AP}_{r}$} & \textbf{$\text{AP}_{c}$} & \textbf{$\text{AP}_{f}$}  \\
\midrule
Mask R-CNN~\cite{gupta2019lvis} & -- &  22.58 & 12.30 & 21.28 & 28.55 \\
\midrule
\multirow{2}{*}{w/ \ourmethod} & First & 23.35  & 16.67 & 21.78  & 28.04 \\
& Second & \textbf{24.28} & \textbf{17.68} & \textbf{22.79} & \textbf{28.83}  \\
\bottomrule
\end{tabular}
\end{table}

\mypara{Comparison with self-training.}
We also study different ways to generate pseudo-masks for training instance segmentation models. We replace \ourmethod segments with those generated by Mask R-CNN (pre-trained on LVIS) ---  treating it as the teacher model to generate pseudo-labels for self-training~\cite{zoph2020rethinking}. We only keep masks whose class labels matched the object-centric image labels to filter out noises. \autoref{tab:supp_self_training} shows the results. All methods use the same training pipeline. \ourmethod outperforms this baseline. We attribute this to the benefit of co-segmentation which explores the similarity across images.

\begin{table}[ht]
\small
\tabcolsep 6pt
\centering
\caption{\small \textbf{Comparison with self-training.}}
\label{tab:supp_self_training}
\begin{tabular}{lcccc}
\toprule
\multicolumn{1}{c}{\textbf{Method}} &  \textbf{AP} & \textbf{$\text{AP}_{r}$} & \textbf{$\text{AP}_{c}$} & \textbf{$\text{AP}_{f}$}  \\
\midrule
MosaicOS~\cite{zhang2021simple} &  24.45 & 18.17 & 23.00 & 28.83 \\
\midrule
w/ LVIS Network & 24.70  & 18.96 & 23.42  & 28.64 \\
w/ \ourmethod & \textbf{25.19} & \textbf{20.23} & \textbf{23.80} & \textbf{28.92}  \\
\bottomrule
\end{tabular}
\end{table}

\mypara{Extended training of \ourmethod.}
Finally, to further compare to Seesaw~\cite{wang2020seesaw}, which applies $2\times$ training scheduling (cf. Section 4.2 of the main paper), we double the training epochs of \ourmethod. \autoref{tab:supp_stronger_schedule} summarizes the results. \ourmethod (2$\times$) achieves further gains and outperforms Seesaw (2$\times$) on all metrics except $\text{AP}_{f}$ for frequent classes. The improvement on $\text{AP}_{r}$/$\text{AP}_{c}$ (\ie, rare/common) is significant, justifying the effectiveness of our approach.

\begin{table}[ht]
\small
\tabcolsep 6pt
\centering
\caption{\small \textbf{\ourmethod with a stronger training schedule.}}
\label{tab:supp_stronger_schedule}
\begin{tabular}{lccccc}
\toprule
\multicolumn{1}{c}{\textbf{Method}} & \textbf{Schedule} & \textbf{AP} & \textbf{$\text{AP}_{r}$} & \textbf{$\text{AP}_{c}$} & \textbf{$\text{AP}_{f}$}  \\
\midrule
Seesaw~\cite{wang2020seesaw} &  2$\times$ & 26.40 & 19.60 & 26.10 & \textbf{29.80} \\
\midrule
MosaicOS~\cite{zhang2021simple} & 1$\times$& 24.45 & 18.17 & 23.00 & 28.83  \\
\multirow{2}{*}{w/ \ourmethod} &  1$\times$& 25.19  & 20.23 & 23.80  &  28.92\\
 & 2$\times$ & \textbf{26.80} & \textbf{21.70} & \textbf{26.90} & 28.60  \\
\bottomrule
\end{tabular}
\end{table}

\section{Qualitative Results}
\label{supp_qual}
One common problem for long-tailed instance segmentation is the trained detector will be overconfident on the frequent objects and suppress the rare objects. \autoref{fig:supp_qual} shows qualitative results. The baseline model tends to predict many false positives which their classes appear more frequently in
the training data. \ourmethod uses augmented training data
with high-quality segments to improve the features, especially for rare objects.

\begin{figure*}[t]
    \centerline{\includegraphics[width=1\linewidth]{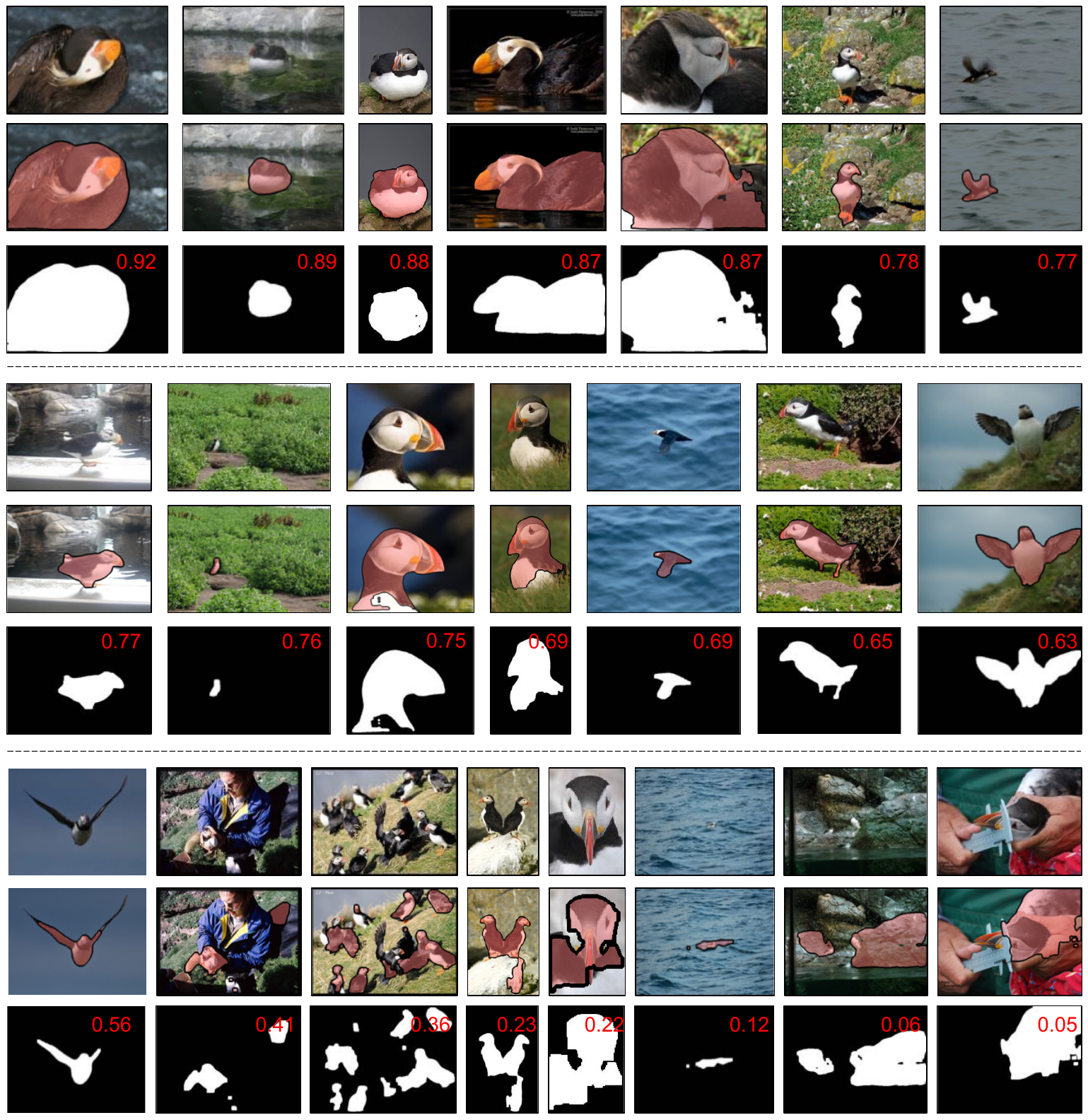}}
    \caption{\small \textbf{Randomly sampled examples of object segments on ImageNet images by \ourmethodbf.} We show a rare class \emph{puffin} in LVIS v1. For each triplet, we show the original image, the object segment, and the binary mask. \ourmethod scores are on the upper right corner of the images. We keep the segments with \ourmethod scores larger than 0.5 (cf. Section 3.2 of the main paper).}
    \label{fig:supp_segment1}
\end{figure*}

\begin{figure*}[t]
    \centerline{\includegraphics[width=1\linewidth]{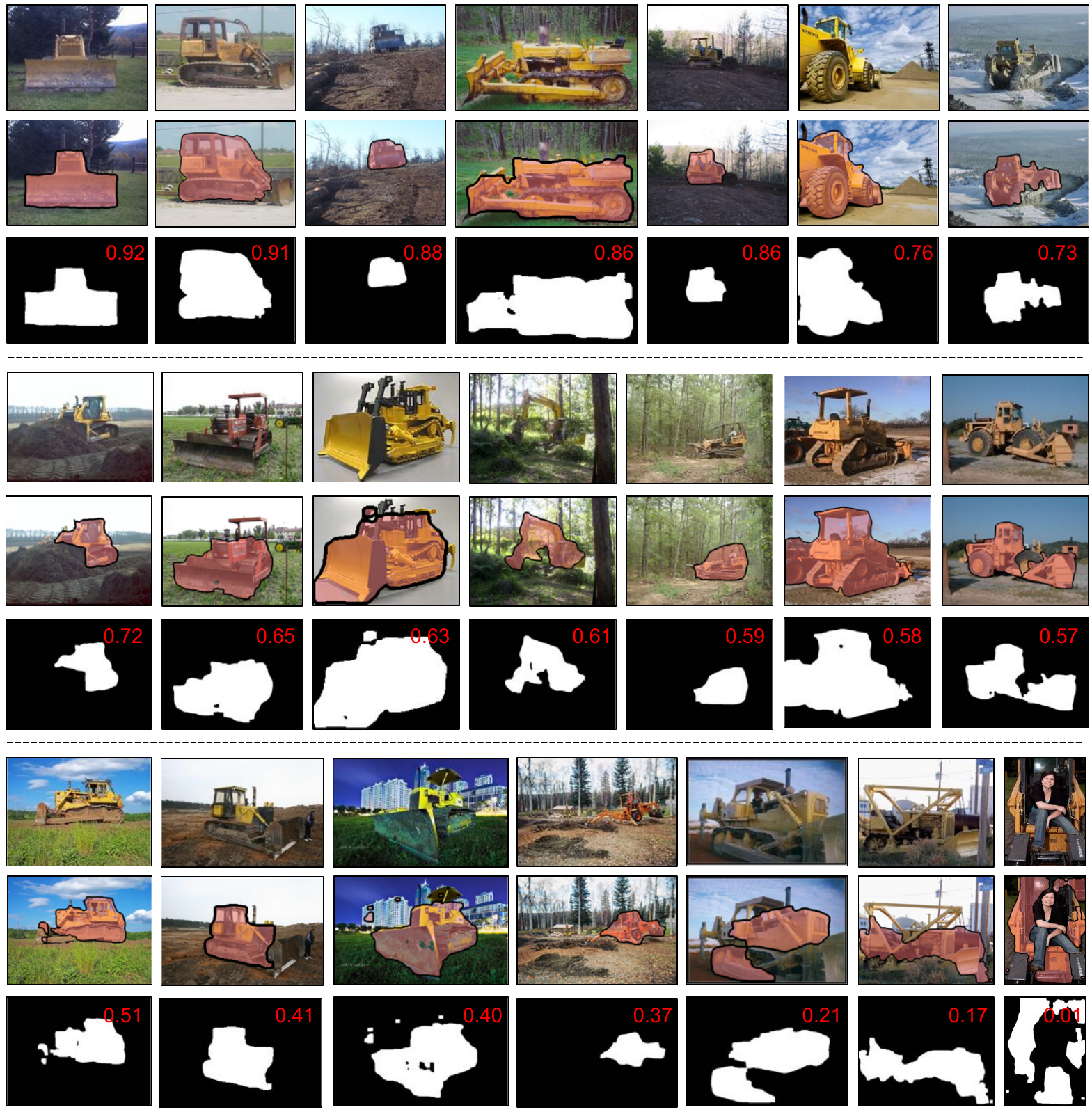}}
    \caption{\small \textbf{Randomly sampled examples of object segments on ImageNet images by \ourmethodbf.} We show a rare class \emph{bulldoze} in LVIS v1. For each triplet, we show the original image, the object segment, and the binary mask. \ourmethod scores are on the upper right corner of the images. We keep the segments with \ourmethod scores larger than 0.5 (cf. Section 3.2 of the main paper).}
    \label{fig:supp_segment2}
\end{figure*}

\begin{figure*}[t]
    \centerline{\includegraphics[width=1\linewidth]{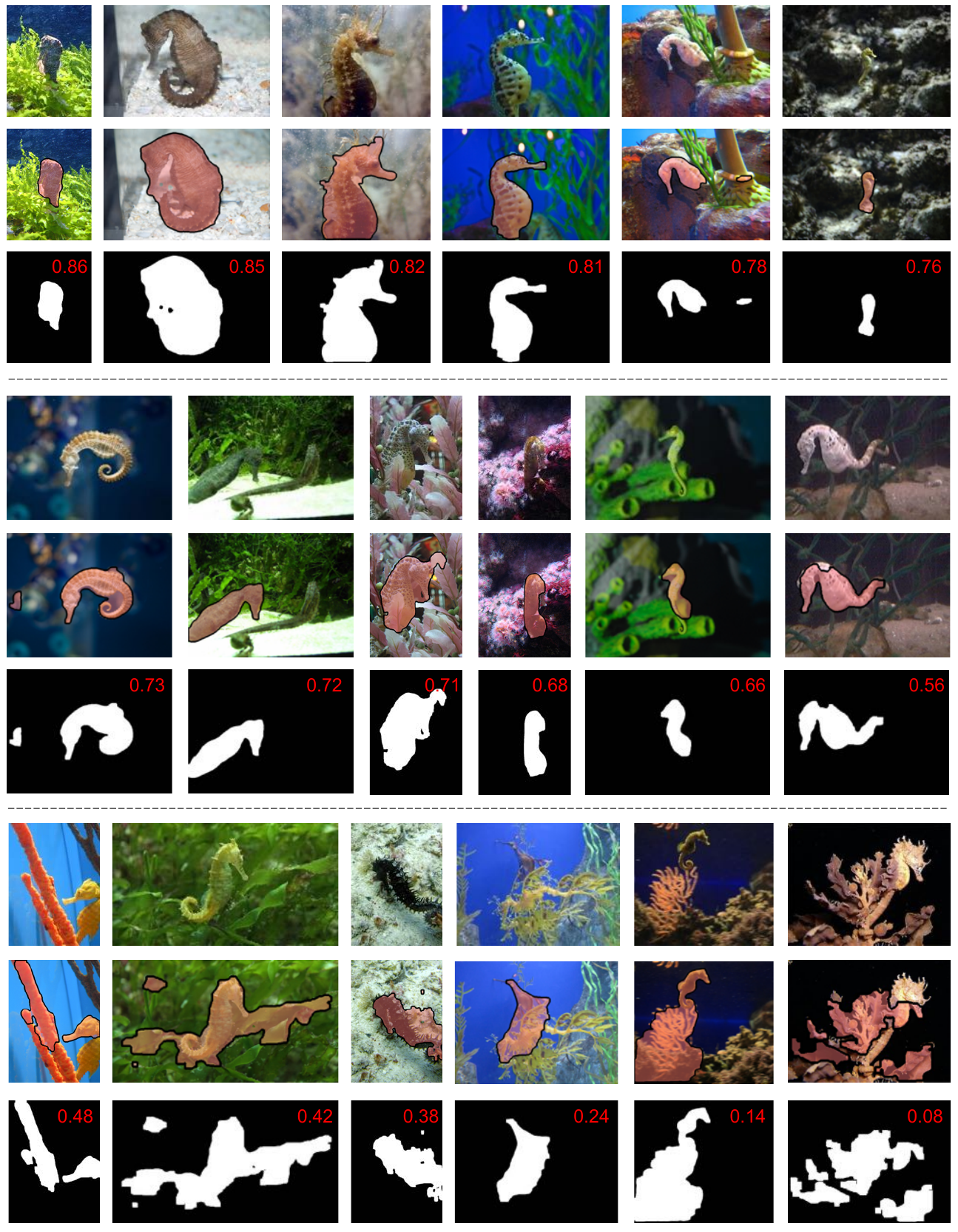}}
    \caption{\small \textbf{Randomly sampled examples of object segments on ImageNet images by \ourmethodbf.} We show a rare class \emph{seehorse} in LVIS v1. For each triplet, we show the original image, the object segment, and the binary mask. \ourmethod scores are on the upper right corner of the images. We keep the segments with \ourmethod scores larger than 0.5 (cf. Section 3.2 of the main paper).}
    \label{fig:supp_segment3}
\end{figure*}

\begin{figure*}[t]
    \centerline{\includegraphics[width=1\linewidth]{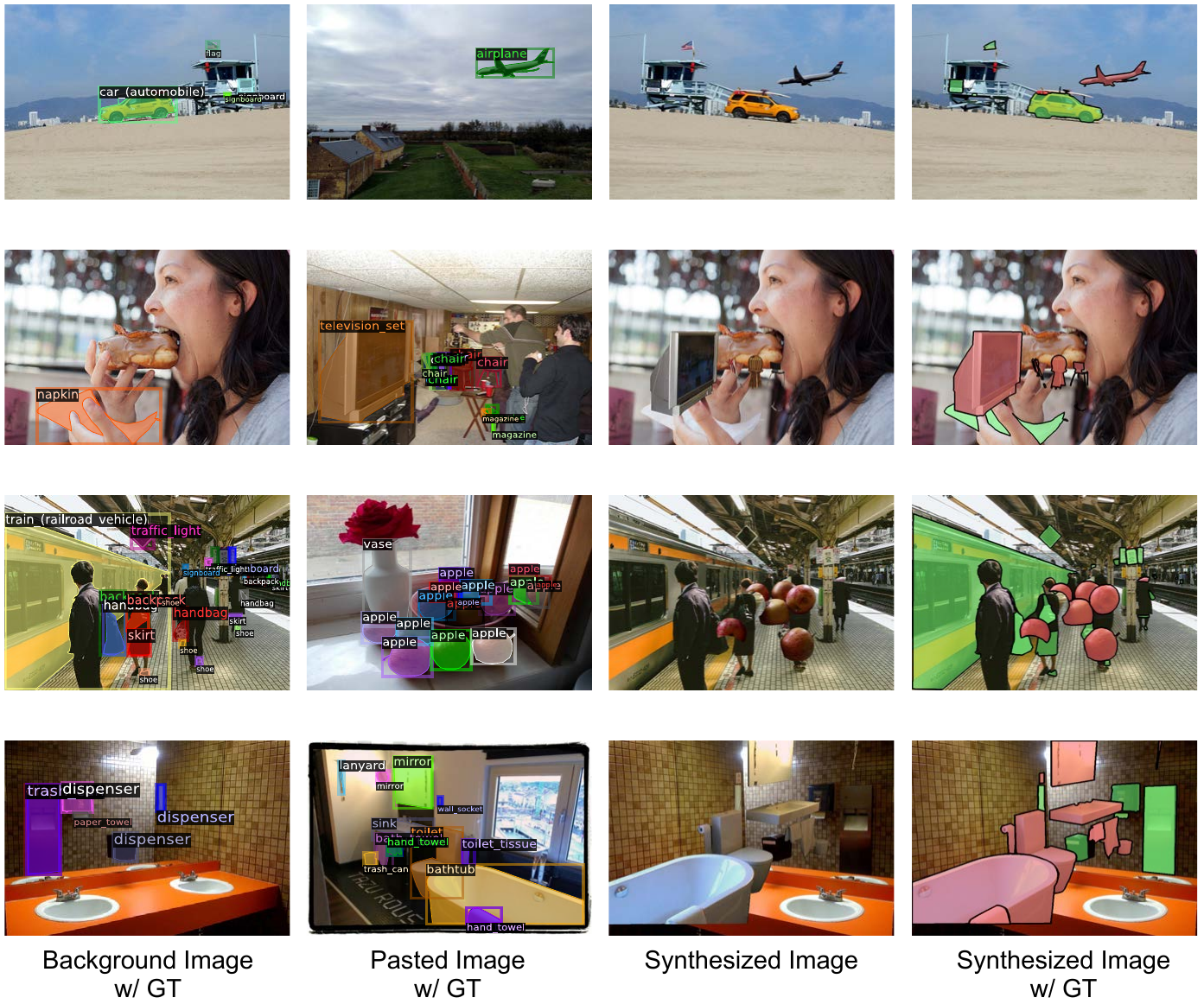}}
    \caption{\small \textbf{Four examples of vanilla copy-paste augmentation using original training images.} For each example, we show the background image with ground-truths, the pasted image with ground-truths, the synthesized image, and the synthesized image with ground-truths. We first randomly pick the background and pasted images from LVIS training set, followed by random shortest edge resize and horizontal flip (cf. Section 4.1 of the main paper). We then select a random number of objects from the pasted image and paste them onto the background image. In the last column, {\color{red}{red}} masks indicate pasted segments; {\color{green}{green}} masks indicate the objects in background images.}
    \label{fig:supp_copy_paste}
\end{figure*}

\begin{figure*}[t]
    \centerline{\includegraphics[width=1\linewidth]{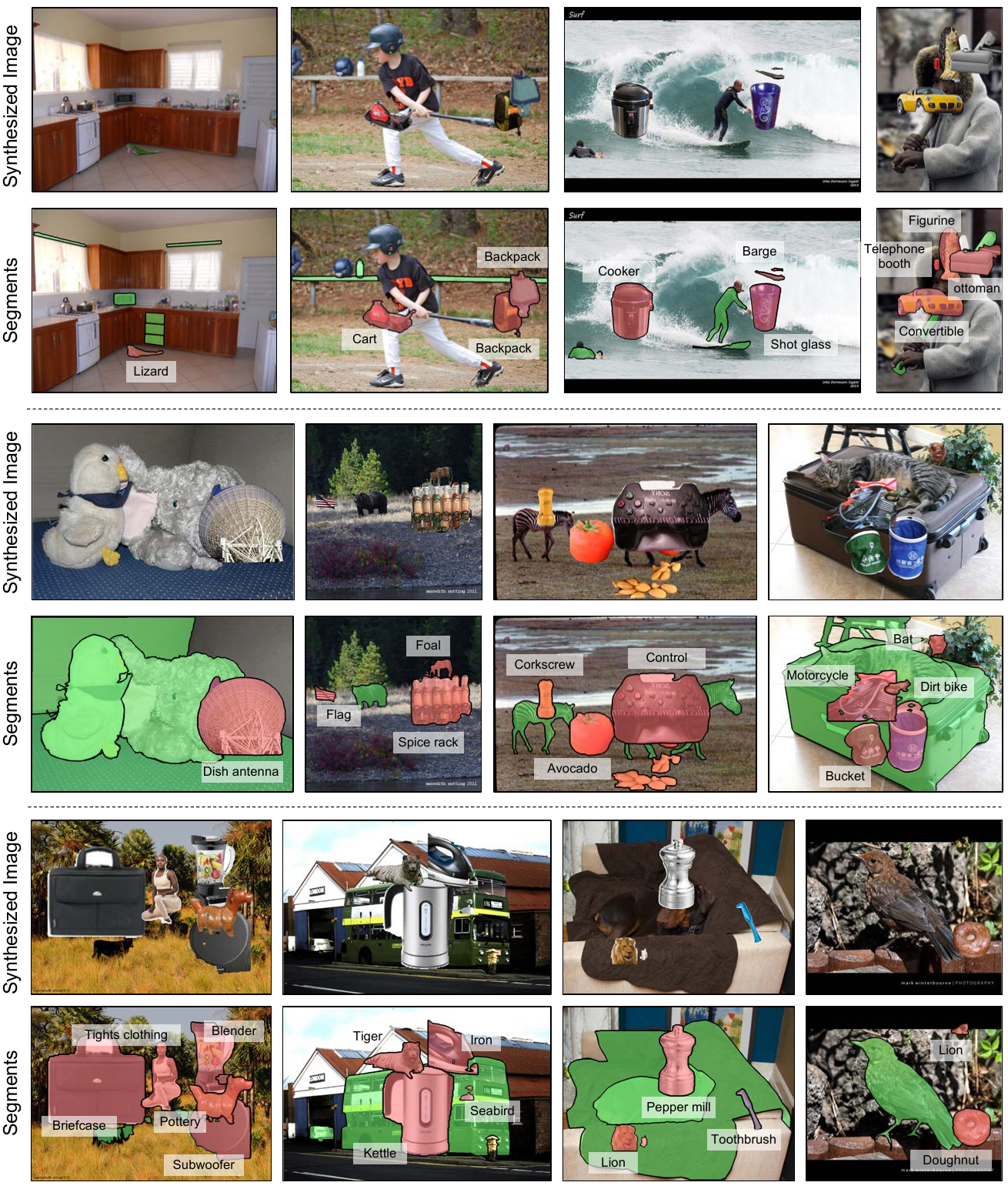}}
    \caption{\small \textbf{Examples of copy-paste augmentation with \ourmethodbf segments.} We generate object segments from object-centric images and randomly paste them onto scene-centric images. {\color{red}{Red}} masks indicate pasted segments by \ourmethod; {\color{green}{green}} masks indicate original objects in scene-centric images.}
    \label{fig:supp_copy_paste_freeseg}
\end{figure*}

\begin{figure*}[t]
    \centerline{\includegraphics[width=1\linewidth]{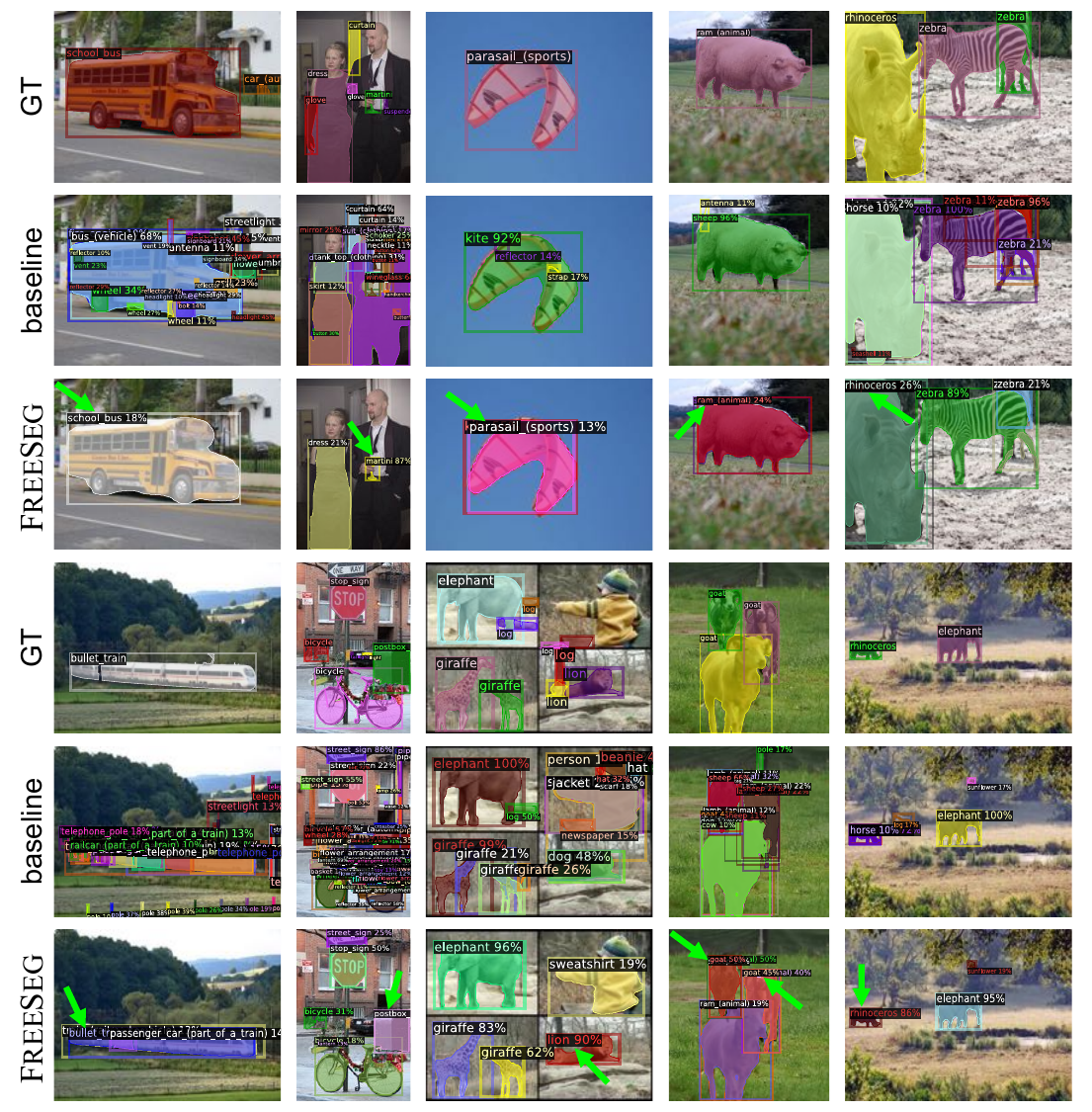}}
    \caption{\small \textbf{Qualitative results. }{\color{green}{Green}} arrows are used to indicate the improvement. \ourmethod successfully detects \textit{school bus}, \textit{martini}, \textit{parasail}, \textit{ram}, \textit{rhinoceros}, \textit{bullet train}, \textit{postbox}, \textit{lion}, and \textit{goat}.}
    \label{fig:supp_qual}
\end{figure*}

\end{document}